\pdfoutput=1
\PassOptionsToPackage{numbers, compress}{natbib}

\documentclass[runningheads]{llncs}
\usepackage{amsmath,pifont}
\usepackage{algorithm}
\usepackage{multicol}
\usepackage{amstext}
\usepackage{multirow}
\usepackage{booktabs}
\usepackage[skip=0pt]{subcaption}
\usepackage{times}
\usepackage{lipsum}
\usepackage[shortlabels]{enumitem}
\usepackage{wrapfig}
\usepackage{array}
\usepackage{siunitx}
\usepackage{csvsimple}
\usepackage[multidot]{grffile}
\usepackage{bbm}
\usepackage{dblfloatfix}
\usepackage{hyperref}
\usepackage{makecell}
\usepackage{bbm, dsfont}
\usepackage{mathtools}
\usepackage{xcolor}
\usepackage{comment}
\usepackage[multiple]{footmisc}
\usepackage{mathrsfs}
  \usepackage{amssymb}
\usepackage{tikz}
\usepackage[underline=false]{pgf-umlsd}
\usetikzlibrary{calc,positioning}

\usepackage{algorithmicx}
\usepackage{algpseudocode}

\usepackage{listings}
\usepackage{ulem}

\renewcommand{\epsilon}{\varepsilon}

\hypersetup{final}

\newcommand{\boldg}{\ensuremath{\boldsymbol{g}}}
\newcommand{\boldh}{\ensuremath{\boldsymbol{h}}}

\newcommand{\boldx}{\bfx}
\newcommand{\boldy}{\ensuremath{\boldsymbol{y}}}
\newcommand{\boldz}{\ensuremath{\boldsymbol{z}}}

\newcommand{\boldI}{\ensuremath{\boldsymbol{I}}}

\newcommand{\boldL}{\ensuremath{\boldsymbol{L}}}

\newcommand{\boldO}{\ensuremath{\boldsymbol{O}}}

\newcommand{\boldR}{\ensuremath{\boldsymbol{R}}}

\newcommand{\boldW}{\ensuremath{\boldsymbol{W}}}

\newcommand{\bfx}{\ensuremath{\mathbf{x}}}

\renewcommand{\Pr}{\mathop{\mathbf{Pr}}}

\makeatletter
\newcommand{\vast}{\bBigg@{4}}
\newcommand{\Vast}{\bBigg@{5}}
\makeatother

\algnewcommand\algorithmicinput{\textbf{Input:}}
\algnewcommand\INPUT{\item[\algorithmicinput]}
\algnewcommand\algorithmicoutput{\textbf{Output:}}
\algnewcommand\OUTPUT{\item[\algorithmicoutput]}
\algnewcommand\algorithmicforeach{\textbf{foreach}}
\algdef{S}[FOR]{ForEach}[1]{\algorithmicforeach\ #1\ \algorithmicdo}
\algnewcommand\algorithmicparallelforeach{\textbf{parallel foreach}}
\algdef{S}[FOR]{PForEach}[1]{\algorithmicparallelforeach\ #1\ \algorithmicdo}

\tikzset{%
  every neuron/.style={
    circle,
    draw,
    minimum size=1cm
  },
  neuron missing/.style={
    draw=none, 
    scale=4,
    text height=0.333cm,
    execute at begin node=\color{black}$\vdots$
  },
}
\usepackage{booktabs}

\newif\ifcomments
\commentstrue

\begin{document}
\title{Scaling Private Deep Learning \\ with Low-Rank and Sparse Gradients}
\author{Ryuichi Ito \inst{1}
\and Seng Pei Liew \inst{2}
\and Tsubasa Takahashi \inst{2}
\and Yuya Sasaki \inst{1}
\and Makoto Onizuka \inst{1}}
\authorrunning{R. Ito, S. Liew et al.}
\institute{Osaka University \email{\{ito.ryuichi, sasaki, onizuka\}@ist.osaka-u.ac.jp}
\and LINE Corporation \email{\{sengpei.liew, tsubasa.takahashi\}@linecorp.com}}
\maketitle              %
\begin{abstract}
Applying Differentially Private Stochastic Gradient Descent (DPSGD) to training modern, large-scale neural networks such as transformer-based models is a challenging task, as the magnitude of noise added to the gradients at each iteration scales with model dimension, hindering the learning capability significantly.
We propose a unified framework, \textsf{LSG}, that fully exploits the low-rank and sparse structure of neural networks to reduce the dimension of gradient updates, and hence alleviate the negative impacts of DPSGD.
The gradient updates are first approximated with a pair of low-rank matrices.
Then, a novel strategy is utilized to sparsify the gradients, resulting in low-dimensional, less noisy updates that are yet capable of retaining the performance of neural networks.
Empirical evaluation on natural language processing and computer vision tasks shows that our method outperforms other state-of-the-art baselines.

\end{abstract}
\section{Introduction}
It is ubiquitous nowadays for technology companies to collect user data to provide better services.
Many of such modern services are empowered by sophisticated deep learning technologies, which excel at various tasks, sometimes outperforming human counterparts.
However, there is a growing privacy concern around these technologies, as it has been shown that deep learning models are vulnerable to privacy attacks \cite{shokri2017membership}, and tend to memorize sensitive training data \cite{carlini2019secret}.
As a result, training deep neural networks (NNs) with rigorous guarantees of privacy has attracted considerable attention lately.

Differential privacy (DP) is a paradigm that aims to expose minimal information about the presence of a particular data instance in a database \cite{dwork2006calibrating,dp,Dwork14}.
In deep learning, the common way of achieving DP is to update the NN parameters with noisy gradients, which is known as Differentially Private Stochastic Gradient Descent (DPSGD) \cite{dpsgd}.
Empirically, NN models trained with DP are shown to be robust to membership inference \cite{nasr2021adversary} and data reconstruction attacks \cite{guo2022bounding}.

While DPSGD is a generic and powerful technique capable of training various NN models, it suffers from utility loss with respect to the scale of NNs.
This can be understood by the following two steps, which compose DPSGD.
\begin{itemize}
    \item \textbf{Gradient clipping.} DPSGD first clips the gradient of each sample (per-sample gradient) $\boldg$ in $l_2$ norm, i.e., $\boldg/\text{max}(\boldg,C)$, where $C$ is the clipping norm. 
    This is to fix the sensitivity of the gradients to be $C$.
    \item
    \textbf{Gaussian mechanism.} Then, the clipped gradients are summed, and isotropic Gaussian noise $z\sim \mathcal{N}(0,C^2\sigma^2)$ is added to \textit{each} dimension of the gradients. Here, $\sigma$ is the noise multiplier determined by the privacy budget $(\epsilon,\delta)$, the number of training iteration $s$, sampling rate $q=B/N$ with $B$ the batch size and $N$ the total number of samples.
\end{itemize}
It can therefore be seen that DPSGD suffers from unfavorable \textit{clipping effects}. Useful information from $\boldg$ is ``clipped'' away when fixing $C$ while increasing the number of NN parameters, as the $l_2$ norm of $\boldg$ increases with it.
Furthermore, the total amount of noise is proportional to the dimension of $\boldg$ (the number of NN parameters).
The clipped and noisy updates make privately trained large-scale NNs low-utility in general.

In this work, we aim to alleviate the shortcomings of DPSGD when used to train large-scale NNs.
To achieve this, we seek to reduce the number of training parameters using the fact that the NN weight matrices are intrinsically redundant.

\noindent \textbf{Our contributions are summarized as follows.}
\begin{itemize}
    \item First, we show that applying either low-rank or sparse assumption of NNs independently as introduced in previous works to training NNs with DP results in sub-par performance~\cite{rgp,snfdpsgd}.
    \item Second, we instead propose a unified strategy, ``\textbf{LSG}" that exploits \textbf{L}ow-rank and \textbf{S}parse properties of NNs' \textbf{G}radient updates. 
    Initially, the gradient updates are approximated with a pair of low-rank matrices. Then, a novel strategy is utilized to sparsify the gradients of the low-rank matrices to minimize the negative impacts of DPSGD.
    \item Lastly, a thorough empirical investigation of natural language processing (NLP) and computer vision (CV) tasks using large-scale neural networks is performed to confirm the efficacy of our proposal. Experimental results show that our method improves up to 40\% compared with sparsification-only methods and 4\% compared with low-rank-decomposition-only methods on various network architectures and learning approaches.
\end{itemize}

The rest of the paper is organized as follows.
A review on DPSGD and related work is given in Section \ref{sec:pre}.
Then, we give empirical observation motivating our approach in Section \ref{subsec:emp}.
Our strategy is described in Section \ref{subsec:exploit} and the full procedure is given in Section \ref{subsec:full_alg}.
In Section \ref{sec:exp}, experimental details are provided.
Before concluding, in Section \ref{subsec:negative}, we also provide negative results that have not led to improvement in performance.

\section{Preliminaries}
\label{sec:pre}
In this section, we first give definitions and known theorems essential for understanding DPSGD and our proposal.
Then, we give a short survey of works related to ours.
We begin with the definition of DP.

\begin{definition}[($\epsilon$, $\delta$)-Differential Privacy \cite{Dwork14}]
Given privacy parameters $\epsilon \geq 0$ and $\delta$ satisfying $1 \geq \delta \geq 0$, a randomized mechanism, $\mathcal{M}: \mathcal{D} \rightarrow \mathcal{S}$ with domain $\mathcal{D}$ and range $\mathcal{S}$ %
satisfies ($\epsilon$, $\delta$)-differential privacy (DP) if for any two adjacent inputs $D, D' \in \mathcal{D}$ and for any subset of outputs $S \subseteq \mathcal{S}$, the following holds:
\begin{align}
\Pr[\mathcal{M}(D) \in S] \leq e^\epsilon \cdot \Pr[\mathcal{M}(D') \in S] + \delta.
\end{align}
\end{definition}
Here, ``adjacent'' refers to $D$ and $D'$ differing in one element.
As in DPSGD, Gaussian mechanism is applied by adding noises to the gradient updates to sanitize the model with DP:
\begin{definition} [Gaussian Mechanism \cite{Dwork14}]
Let $f: X \rightarrow \mathbb{R}$ be an arbitrary function with sensitivity $\Delta_{f}$.
The Gaussian Mechanism, $\mathcal{M}_\sigma$, parameterized by $\sigma$, adds noise to the output of $f$ as follows:
\begin{equation}
\label{defn:gaussian_mechanism}
  \mathcal{M_\sigma}(x) = f(x) + \mathcal{N}(0,\sigma^2 I).
\end{equation}
\end{definition}
The following theorem ensures that DP-sanitized quantities can be reused without further spending privacy budgets:
\begin{theorem}[Post-processing Theorem \cite{Dwork14}] 
\label{thm:post}
Let $\mathcal{M}: \mathcal{D} \rightarrow \mathcal{R}$ be ($\epsilon,\delta$)-DP and let $f:\mathcal{R}\rightarrow \mathcal{R'}$ be an arbitrary function. Then, $f \circ \mathcal{M}: \mathcal{D} \rightarrow \mathcal{R'}$ is ($\epsilon,\delta$)-DP.
\end{theorem}

\noindent\textbf{Privacy accounting.}
As NNs are typically updated for multiple iterations, one needs to account for the total privacy budget spent, as each iteration exposes some information of the data, inducing privacy losses.
The original DPSGD work introduces moments accountant to take care of it \cite{dpsgd}.
We use the more recent numerical accountant for calculating the total privacy budget, as it leads to a much tighter privacy accounting \cite{gopi2021numerical}.

\noindent\textbf{Notations.}
Some commonly used notations are introduced here.
In NNs, the output response of input $\boldx$ and output $\boldy$ of a fully connected layer is related to its weight matrix $\boldW \in \mathbb{R}^{m \times n}$ as follows:
\begin{equation*}
    \boldy = \boldW^T \boldx,
\end{equation*}
where $\boldx \in \mathbb{R}^{m}$ is an $m$-dimensional input vector and $\boldy \in \mathbb{R}^{n}$ is an $n$-dimensional output vector. 
We use bold text to represent vectors or tensors.
$W_{ij}$ denotes the elements of $\boldW$ for $i \in [m]$ and $j \in [n]$.
Slightly abusing notations, we use asterisk to indicate all elements of a certain set.
That is, using the above example, $W_{i*}$ means $W_{ij}$ for all $j \in [n]$.
$r$ commonly refers to the rank of a matrix.
When we say that we are performing $p\%$ sparsification on a certain matrix, it means that elements other than the top largest $(100-p)\%$ elements in value are made zero.

\subsection{Related Work}
\noindent \textbf{Beyond DPSGD.}
Early successes of DPSGD were limited to training relatively simple convolutional NNs on small dataset such as MNIST \cite{dpsgd}.
Since then, various studies have been conducted to overcome the limitations of DPSGD with respect to large-scale NNs \cite{yu2021not,snfdpsgd,rgp}.
In particular, a method called reparametrized gradient perturbation (RGP) introduced recently utilizes the low-rank properties in the gradient updates to enhance the performance of DPSGD \cite{rgp}.
As will be shown later, our work builds on RGP by exploiting more thoroughly the redundant structure of the gradient updates.

\noindent \textbf{Network pruning.}
Pruning techniques have been utilized to reduce computation and storage costs of NNs \cite{pruning_synapses,pruning_conv,pruning_any,pruning_units}.
Typically, NNs weights small in magnitude are pruned and empirically they do not affect the final performance much.
Inspired by these works, \cite{snfdpsgd} proposes to freeze (stop updating) unimportant weights while training with DPSGD to reduce the total amount of noise injected to the updates.
However, as will be elaborated in the next sections, the proposal's performance is less-than-satisfactory. 
Instead, we demonstrate that our approach is more effective performance-wise.

In summary, previous works on improving DPSGD neglect either the low-rankness or the sparsity of NNs, compromising the performance of NNs.
Our purpose is to seek a unified framework integrating these two features to gain further improvements.

\section{Private Learning with Low-rank and Sparse Gradients (LSG)}
In this section, we propose our method of training large-scale NNs with DP.
Before doing so, we provide empirical observation that inspires our approach.
Subsequently, the full procedure and algorithm are described in detail.

\subsection{Empirical Observations on the NN Weight Matrices}
\label{subsec:emp}
\begin{figure}
     \centering
     \begin{subfigure}[b]{0.4\textwidth}
         \centering
         \includegraphics[width=0.92\hsize]{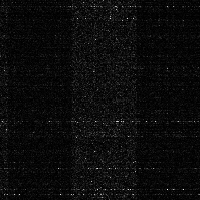}
                           \caption{Weight matrix}
                \label{fig:subexp_roberta_w}
     \end{subfigure}
     \begin{subfigure}[b]{0.4\textwidth}
         \centering
         \includegraphics[width=0.92\hsize]{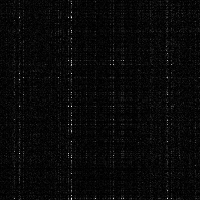}
                  \caption{Gradient matrix}
                \label{fig:subexp_roberta_g}
     \end{subfigure}
     \caption{Intensity plots of NN weights and gradients.
     Here, the plotted weights are from the first fully connected layers of the pretrained RoBERTa.
     The inputs (outputs) are in the horizontal (vertical) direction.
     The gradients are from the corresponding layer obtained by running a batch of data taken from SST-2.
     }
     \label{fig:subexp_roberta}
\end{figure}

It is well known that neurons in a NN do not contribute equally and NNs are intrinsically overparametrized.
Such redundant structure has been used to, e.g., compress NNs  \cite{powersgd,compressionbylowrankapprox,compressionbysparsityandlowrankapprox1,compressionbysparsityandlowrankapprox2}.
To motivate our approach, let us, as an example, look at a layer of RoBERTa \cite{liu2019roberta}, a pretrained transformer model.
We make a 2D plot of the absolute values of $|W_{ij}|$ as well as the corresponding gradients, $|\partial W_{ij}|$ in Figure \ref{fig:subexp_roberta}.

Based on the figure, one can make two important empirical observations.
First, the distribution is sparse: there exist only a limited number of gradient updates where the absolute values are significantly large.
This indicates that it is enough to perform updates on only a portion of the parameters.
Second, weights measured in the unit of \textit{neurons} are dense, as can be observed from the contrastive horizontal/vertical lines in the figure.
That is, for certain fixed $i$ ($j$) of $W_{ij}$, all $j\in [n]$ ($i\in [m]$) of $W_{ij}$ are dense.
We will leverage these observations to craft our approach in the following.

\subsection{Redundant Structure of NNs}
To enable the training of large-scale NNs with DP, we reduce the number of training parameters, which in turn mitigate the clipping effects and also reduce the total noises added to the gradients.

We attack the problem via two perspectives of redundancy in NNs: low-rank approximation and sparsification.
Note that low-rank approximation is commonly performed using the singular value decomposition (SVD) method, which decomposes the matrix via a set of orthonormal bases.
This may be thought as a sparse approximation at the ``correct" bases of weight matrix.
In this sense, low-rankness and sparsity are closely related and can be complementary to each other.

\noindent \textbf{Low-rank approximation.}
Let us first give a quantitative description of the efficacy of low-rank approximations in reducing the number of training parameters.
The weight matrix is approximated by a pair of matrix of rank $r$, $\boldW \simeq \boldL \boldR \; (\boldL\in \mathbb{R}^{m \times r}, \boldR\in \mathbb{R}^{r \times n})$. $\boldL$ and $\boldR$ contain $r(m + n)$ parameters in total, in comparison with the number of parameters of $\boldW$, which is $m \times n$.
By choosing $r\ll \text{min}(m,n)$, one is able to reduce the number of training parameters drastically: $r(m + n) \ll m \times n$.
We use the terms approximation and decomposition interchangeably to describe the procedure of writing $\boldW$ as $\boldL \boldR$.

\noindent \textbf{Sparsification.}
Network pruning techniques commonly prune neurons with small absolute weights to reduce the size of NNs.
For our purposes, instead of pruning, we freeze/mask out the corresponding gradient updates of these weights.
This reduce the number of training parameters, and alleviate the clipping effects as well as reduce the total noise due to DPSGD.
The straightforward way of sparsification is choosing the top $W_{ij}$ for $i\in [m]$, $j\in [n]$ with respect to the magnitude of the matrix weights to perform updates ($p\%$ sparsification)  \cite{pruning_synapses}. 
This approach is also known to be effective even when grouped by structural units such as \textit{kernel} \cite{pruning_conv} in convolutional layers and \textit{neural node} \cite{pruning_any,pruning_units}.

We note that, instead of pruning the sparsified weights (as done in the network pruning literature), we retain the full network structure during inference to ensure minimal utility loss due to pruning (as it is not our main purpose to reduce the model size via pruning).

\subsection{Exploiting the Redundant Structure of NNs with LSG}
\label{subsec:exploit}
Here, we describe our proposal, \textsf{LSG} that combines the two orthogonal approaches of low-rank approximation and sparsification.
As they cannot be simply applied together, in the following, we present our proposal that gets the best out of both worlds to achieve state-of-the-art results.

We first perform low-rank approximation on the NN weight matrices ($\boldW = \boldL \boldR$) with a pair of matrices $\boldL,\boldR$ of rank $r$.
Under this approximation, we update the parameters of $\boldL$ and $\boldR$.
They are related to the gradient updates of the original matrix $\partial\boldW$ as follows \cite{rgp}:
\begin{eqnarray}
\partial{\boldW} = (\partial{\boldL})\boldR + \boldL(\partial{\boldR}) - \boldL\boldL^{T}(\partial{\boldL})\boldR
\label{eq:rgp}
\end{eqnarray}
Note that the updates $\partial{\boldL}$ and $\partial{\boldR}$ are to be DP-sanitized with the Gaussian mechanism.
Sanitizing $\boldL$ and $\boldR$ is not necessary as they are derived from $\boldW$, which is DP by post-processing theorem.

Updating with Equation \ref{eq:rgp} does not exploit the sparsity of NN weight matrices.
To do so, as we are updating $\boldL$ and $\boldR$, we propose to freeze the unimportant gradients in $\partial{\boldL}$ and $\partial{\boldR}$.
Note that this is nontrivial as we are working on $\partial{\boldL}$ and $\partial{\boldR}$ but not $\partial{\boldW}$; one could simply freeze unimportant $\boldW$ based on its absolute values if working directly on $\boldW$. 
How should we choose the unimportant components of $\partial{\boldL}$ and $\partial{\boldR}$? 

Let us first define the following \textit{importance} metrics based on the observation of contrastive horizontal/vertical lines in Figure \ref{fig:subexp_roberta} to achieve sparsification \cite{compressionbysparsityandlowrankapprox2}: 
\begin{eqnarray}
I_i &=& \sum_{j=1}^{n}|W_{ij}| \label{eq:iimportance} \\
O_j &=& \sum_{i=1}^{m}|W_{ij}| \label{eq:oimportance}
\end{eqnarray}

Recall that under low-rank approximation, $W_{ij} = \sum_{l=1}^r L_{il}R_{lj}$.
Hence, $L_{i*}$ for $i$ where $I_i$ is large correspond to ``important" $L$'s, or important $W_{i*}$.
The same applies to $R_{*j}/W_{*j}/O_j$.

Gaining insights from network pruning techniques that prune unimportant NN weights to save space, we propose to sparsify the gradients of $\boldL$ and $\boldR$ based on the importance metrics (Equation \ref{eq:iimportance} and \ref{eq:oimportance}).
More precisely, we perform the following sparsification:
\begin{eqnarray*}
\forall{j} \in [r], \partial{\boldL} &=& \begin{cases}
\partial{L}_{ij} & \text{if } I_i \text{ in top-$(100-p)\%(\boldI$)} \\
0 & \text{otherwise}
\end{cases}
\\
\forall{i} \in [r], \partial{\boldR} &=& \begin{cases}
\partial{R}_{ij} & \text{if } O_j \text{ in top-$(100-p)\%(\boldO$)} \\
0 & \text{otherwise}
\end{cases}
\end{eqnarray*}
As a result of this sparsification, the number of update parameters is further reduced to $(100-p)r(m + n)/100 \leq r(m + n) \ll m \times n$.

\begin{figure}[ht]
    \centering
    \includegraphics[width=0.6\hsize]{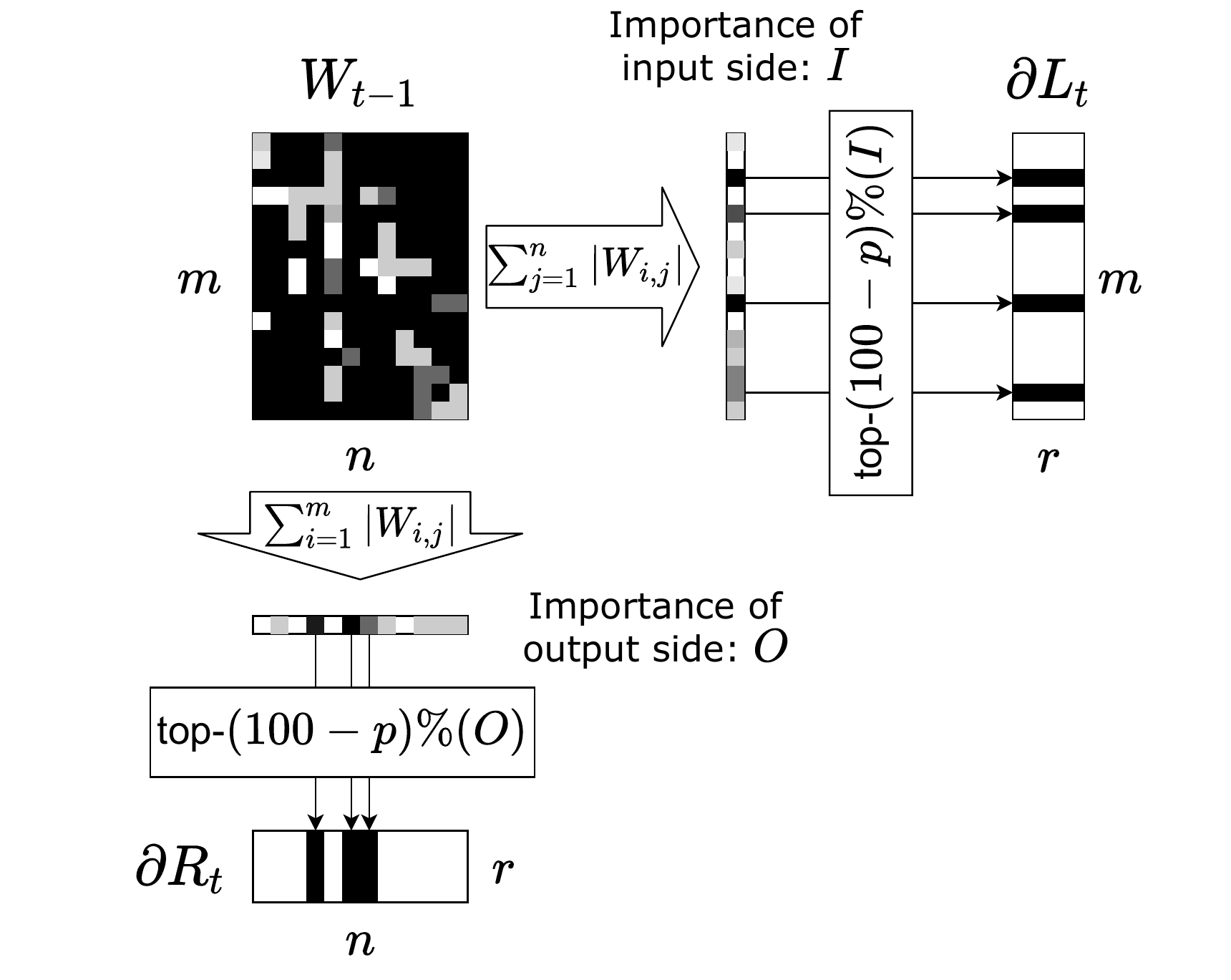}
    \caption{An illustration of sparsification of low-rank gradient matrices of fully connected and attention layers by the weight matrix. Blacked-out elements in $\partial{\boldL}$ and $\partial{\boldR}$ means that they are frozen and will not be updated at time step $t$. \label{fig:prop_detail}}
\end{figure}

Note from Equation \ref{eq:rgp} that $\partial W_{ij} =0$ if and only if $\partial L_{i*} = 0$ and $\partial R_{*j} = 0$.
That is, only the weights of which $I_i$ and $O_j$ are simultaneously small are frozen during update in our approach.
This is sufficient empirically even though not all $W$'s will small absolute values are frozen (as we are updating the low-rank matrices), as the total noise is scaled down by dimension (number of parameters) reduction.
Also note that our sparsification strategy works for any low-rank matrices independent of the underlying approximation method.
An illustration of our proposal is shown in Figure \ref{fig:prop_detail}.

\subsection{Extension of LSG to Convolutional Layers}
\label{subsec:ext_conv}
\begin{figure}[ht]
    \centering
    \includegraphics[width=0.7\hsize]{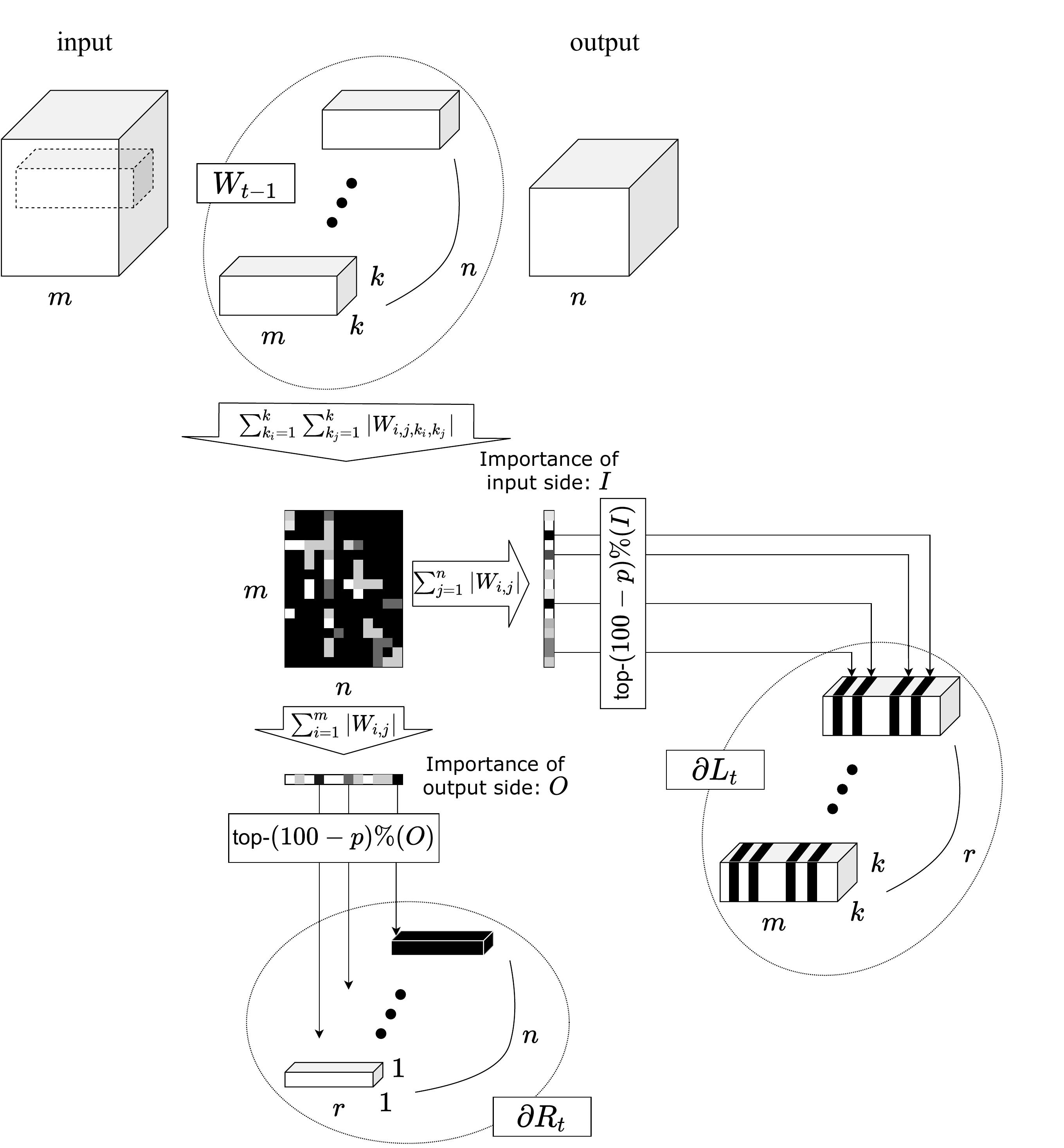}
    \caption{An illustration of sparsification of low-rank gradient matrices of convolutional layers by the weight matrix. Blacked-out elements in $\partial{\boldL}$ and $\partial{\boldR}$ means that they are frozen and will not be updated at time step $t$. \label{fig:prop_conv}}
\end{figure}

The description of our proposal in Section \ref{subsec:exploit} applies in a straightforward way to fully connected and attention layers.
To apply our proposal to convolutional layers, some extensions are required and will be described in the following.
For an illustration, see Figure \ref{fig:prop_conv}.

The convolutional kernel may be described by $\boldW \in \mathbb{R}^{n\times m \times k \times k}$ which maps an input feature $\boldx \in \mathbb{R}^{m\times w \times h}$ to an output feature $\boldh \in \mathbb{R}^{n\times w' \times h'}$.
Fixing a coordinate in the second and third dimension of $\boldh$, i.e., $h_{:,i,j}$, its value is related to a feature $\boldx \in \mathbb{R}^{m\times k \times k}$ and a weight $\tilde{\boldW} \in \mathbb{R}^{n\times mk^2}$ as follows: $\boldh_{:,i,j} = \tilde{\boldW}^T\boldx^{(i,j)}$.
Here, $\tilde{\boldW}$ is $\boldW$ but with the last three dimesions flattened into a single dimension.
In this way, we can write the low-rank approximation as $\tilde{\boldW} = \tilde{\boldL} \tilde{\boldR}$ where $\tilde{\boldL} ,\tilde{\boldR}$ are the flattened versions of low-rank matrices
$\boldL \in \mathbb{R}^{n \times r \times 1 \times 1}, \boldR \in \mathbb{R}^{r \times m \times k \times k}$.

Based on this observation, we define the corresponding importance metrics at the \textit{kernel} level:
\begin{eqnarray}
I_i &=& \sum_{j=1}^{n}\sum_{k_i=1}^{k}\sum_{k_j=1}^{k}|W_{ijk_{i}k_{j}}| \label{eq:conv_iimportance} \\
O_j &=& \sum_{i=1}^{m}\sum_{k_i=1}^{k}\sum_{k_j=1}^{k}|W_{ijk_{i}k_{j}}| \label{eq:conv_oimportance}
\end{eqnarray}

Using these metrics, sparsification can then be performed analogously as follows:
\begin{eqnarray*}
\forall{j} \in [r], \forall{k_{i}} \in [k], \forall{k_{j} \in [k]},
\partial{\boldL} &=& \begin{cases}
  \partial{L}_{ijk_{i}k_{j}} & \text{if } I_i \text{ in top-$(100-p)\%(\boldI$)} \\
  0 & \text{otherwise}
\end{cases}
\\
\forall{i} \in [r], \forall{k_{i}} \in [k], \forall{k_{j} \in [k]},
\partial{\boldR} &=& \begin{cases}
  \partial{R}_{ijk_{i}k_{j}} & \text{if } O_j \text{ in top-$(100-p)\%(\boldO$)} \\
  0 & \text{otherwise}
\end{cases}
\end{eqnarray*}

\subsection{The Full Algorithm of LSG}
\label{subsec:full_alg}
Let us describe the full algorithm based on insights obtained above.
The full procedure is described in Algorithm \ref{alg:prop}.\footnote{Due to space constraint, we show only full algorithms applicable to fully connected and attention layers.
They can be trivially extended to those applicable to the convolutional layer using arguments given in Section \ref{subsec:ext_conv}.}

At the beginning of training, The NN weights are initialized randomly or pretrained with public data.
Note that the end of each iteration of the algorithm, the NN weights will have been DP-sanitized appropriately.
Hence, we can manipulate (perform, e.g., low-rank approximation, sparsification) the weights at will without costing further privacy budget due to the post-processing theorem.

At each iteration, for each layer, we first calculate the importance metrics using Equations \ref{eq:iimportance} and \ref{eq:oimportance}.
Next, we low-rankly approximate the weight matrices.
A single-step power method is used for low-rank approximation, which can be seen as a form of random projection followed by orthonomalization (Algorithm \ref{alg:decompose_pi}).
According to our experimentations, this is more training cost-effective than the usual power method with multiple steps.
Then, a random minibatch of data is sampled, and the sample gradients, $\partial\boldL$, $\partial\boldR$ are calculated from $\partial\boldW$ using the following \cite{rgp}:
\begin{equation}
    \partial \boldL = (\partial \boldW) \boldR^T, \quad \partial \boldR = \boldL^T (\partial \boldW).
    \label{eq:rgp2}
\end{equation}
Subsequently, we perform sparsification on the gradient matrices as in lines \ref{l:sparsifyl} and \ref{l:sparsifyr} of Algorithm \ref{alg:prop}.

Finally, we clip the remaining non-zero gradients, sum them, and add Gaussian noises (only to non-sparse gradients) to achieve DP (Algorithm \ref{alg:dp}).
We update the weights in the form of $\boldW$ using Equation \ref{eq:rgp} and off-the-shelf optimizer (e.g., Adam).
The numerical accountant \cite{gopi2021numerical} is used to track the total privacy budget spent.
The above iteration is stopped once one achieves the desired level of privacy.

\begin{algorithm}[]
\caption{\textsf{LSG} at each step. \label{alg:prop}}
\small
\begin{algorithmic}[1]
\INPUT{Weights at previous step $\boldW_{t-1}^{(l)} \in \mathbb{R}^{m \times n}$ for layers $l \in H$, current step $t$, variance $\sigma^2$, clipping threshold $C$, rank $r$, sparsity $p$, dataset $S$, probability $q$, external low-rank mechanism $Decompose$, DP mechanism $DP$, update mechanism $Update$, numerical accountant~\cite{gopi2021numerical}} 
\OUTPUT{Weights at step $t$ $\boldW_{t}^{(l)}$}

\medskip

\ForEach{$l \in H$}
\State {//\ $\boldW_{t-1}^{(l)}$ is randomly initialized or pre-trained with public datasets or trained with private datasets by DP-SGD}
\ForEach {$i \in [m]$} \label{l:sfor_sumi}
    \State {$I_{i}^{(l)} \gets \sum_{j=1}^{n}|W_{t-1,(i,j)}^{(l)}|$} \label{l:sumi}
\EndFor \label{l:efor_sumi}
\ForEach {$j \in [n]$} \label{l:sfor_sumo}
    \State {$O_{j }^{(l)} \gets \sum_{i=1}^{m}|W_{t-1,(i,j)}^{(l)}|$} \label{l:sumo}
\EndFor \label{l:efor_sumo}
\State{Unimportant input units $\tilde{\boldI}^{(l)} \gets$ top-$(100-p)\%(\boldI^{(l)}$)} \label{l:toppi}
\State{Unimportant output units $\tilde{\boldO}^{(l)} \gets$ top-$(100-p)\%(\boldO^{(l)}$)} \label{l:toppo}

\State{${\boldL}_{t}^{(l)}, {\boldR}_{t}^{(l)} \gets Decompose(\boldW^{(l)}_t, r)$} \Comment{Use low-rankness (Alg. \ref{alg:decompose_pi})} \label{l:lr}
\EndFor
      \State Sample a minibatch $S_q=\{x_q\}_{x_q\in S}$ with probability $q$
      \State Calculate per-sample gradients with Eq \ref{eq:rgp2}
\ForEach{$l \in H$}
\ForEach {$i \in \tilde{\boldI}^{(l)}$} \label{l:sfor_sparsifyl}
    \ForEach{$j \in [r]$}
        \State {$\partial{L}_{t,(i,j)}^{(l)} \gets 0$} \Comment{Use sparsity} \label{l:sparsifyl}
    \EndFor
\EndFor \label{l:efor_sparsifyl}
\ForEach {$j \in \tilde{\boldO}^{(l)}$} \label{l:sfor_sparsifyr}
    \ForEach{$i \in [r]$}
        \State {$\partial{R}_{t,(i,j)}^{(l)} \gets 0$} \Comment{Use sparsity} \label{l:sparsifyr}
    \EndFor
\EndFor \label{l:efor_sparsifyr}
\State{$\tilde{\partial}{\boldL}_{t}^{(l)} \gets DP(\{\partial{\boldL}_{t}^{(l)}(x_q)\}_{q \in S_q}, C, \sigma^2, \tilde{\boldI}^{(l)})$} \Comment{Clip and Add noise (Alg. \ref{alg:dp})} \label{l:dpl}
\State{$\tilde{\partial}{\boldR}_{t}^{(l)} \gets DP(\{\partial{\boldR}_{t}^{(l)}(x_q)\}_{q \in S_q}, C, \sigma^2, \tilde{\boldO}^{(l)})$} \Comment{Clip and Add noise (Alg. \ref{alg:dp})} \label{l:dpr}
\State {$\boldW_{t}^{(l)} \gets Update(\boldW_{t-1}^{(l)}, \tilde{\partial}{\boldL}_{t}^{(l)}, \tilde{\partial}{\boldR}_{t}^{(l)})$
}\label{l:update}
\Comment{Use off-the-shelf optimizer and Eq. \ref{eq:rgp}}
\EndFor
\State {Update numerical accountant}
\State {$\textbf{return} \ \boldW_{t}$} \label{l:return}
\end{algorithmic}
\normalsize
\end{algorithm}

\begin{algorithm}[tb]
   \caption{Low-rank approximation.}
   \label{alg:decompose_pi}
\begin{algorithmic}[1]
    \INPUT{Weight $\Delta$, rank $r$}
    \OUTPUT{Gradient matrices $\boldL\in\mathbb{R}^{m\times r}$,  $\boldR\in\mathbb{R}^{r\times n}$}
  
  \medskip

    \State Initialize  $\boldR$ from standard Gaussian distribution $\mathcal{N}(0,1)$.
  
    \State $\boldL \leftarrow \Delta \boldR^{T}$
    \State Orthonormalize the columns of $\boldL$.
    \State $\boldR \leftarrow \boldL^{T}\Delta$
    \State Orthonormalize the rows of $\boldR$.

    \State Return $\boldL$, $\boldR$
\end{algorithmic}
\end{algorithm}

\begin{algorithm}[tb]
   \caption{Differentially private sanitization.}
   \label{alg:dp}
\begin{algorithmic}[1]
    \INPUT{Per-sample low-rank gradient matrices $\partial\boldL(x_q)\in\mathbb{R}^{m \times r}$ or $\partial\boldR(x_q)\in\mathbb{R}^{r \times n}$ with respect to a minibatch of data $\{x_q\}$, unimportant input or output units $\tilde{\boldI}$ or $\tilde{\boldO}$, variance $\sigma^2$, clipping size $C$}
    \OUTPUT{DP sanitized gradient matrices $\tilde{\partial}\boldL\in\mathbb{R}^{m \times r}$,  $\tilde{\partial}\boldR\in\mathbb{R}^{r \times n}$}
  
  \medskip
\State Clip per-sample gradients with $L_{2}$ norm threshold $C$;
   
    \State Sum per-sample gradients to obtain  $\partial\boldL\;\text{or}\;\partial\boldR$;
\State Perturb with noise $\boldz$ sampled from $\mathcal{N}(0,\sigma^{2}C^{2})$ masked with $\tilde{\boldI}$ or $\tilde{\boldO}$:
       $$\tilde{\partial}\boldL  \leftarrow \partial\boldL + \boldz\; \text{or} \quad \tilde{\partial}\boldR \leftarrow \partial\boldR+\boldz;$$   

    \State Return $\tilde{\partial}\boldL$ or $\tilde{\partial}\boldR$
\end{algorithmic}
\end{algorithm}

\section{Empirical Evaluation}
\label{sec:exp}
To test the efficacy of our proposal, we experiment with NLP tasks using a transformer-based \cite{vaswani2017attention} model (consisting of fully connected and attention layers), and a CV task with Resnet \cite{he2016deep} (consisting mainly of convolutional layers). 

\subsection{Private Fine-tuning of Transformer-based Model}
\label{subsec:nlp}
Recent advances in NLP show that transformer-based large language models (LLMs) pretrained with a large amount of public dataset are capable of reaching state-of-the-art results when fine-tuned to a task-specific dataset \cite{devlin2018bert}.
To leverage the advantage of public (non-private) data, we consider fine-tuning pretrained LLMs with private datasets.

The pretrained model used is RoBERTa-base \cite{roberta}, which contains approximately 1.25 million parameters.
The General Language Understanding Evaluation (GLUE) benchmark \cite{glue}, specifically the SST-2 (sentimental analysis), QNLI (question-answering natural language inference), QQP (semantic similarity of a pair of questions) and MNLI\footnote{We report the average score of two test datasets.} (multi-genre language inference) datasets are used for private fine-tuning.

\noindent\textbf{Baselines.}
We compare our proposal (\textsf{LSG}) with several baselines.
The first is the non-private SGD (\textsf{N.P.}).
The second baseline is the vanilla DPSGD (\textsf{DPSGD}).
The other two baselines are RGP (\textsf{RGP}) and Sparse DPSGD (\textsf{Sparse DPSGD}), corresponding to only performing low-rank approximation/sparsification respectively.
Note also that \textsf{Sparse DPSGD} may be seen as an extension of \cite{snfdpsgd}.
All baselines and our proposal utilize the same network architecture. 

\noindent\textbf{Experimental details.}
Our procedure of low-rank approximation and sparsification is applied to all fully connected and attention layers except the classification layer.
The hyperparameters are set to the following, following \cite{rgp}: Adam optimizer is used, clipping size 10; batch size 2,000; epoch 20; learning rate $10^{-3}$; the privacy parameter $\delta=10^{-5}$ for SST-2/QNLI and $\delta=10^{-6}$ for QQP/MNLI.
The three parameters we are interested in are $\epsilon$, rank $r$ and sparsity $p$.
We scan over these parameters ($\epsilon \in \{1.2, 1.6, 3.3, 6.5\}$ for SST-2/QNLI, $\epsilon \in \{0.9, 1.7, 3.4, 6.7\}$ for QQP/MNLI, $r \in \{2, 4, 8\}$, and $p \in \{0, 0.1, 0.3, 0.5\}$) to compare the performances.
\footnote{More precisely, we vary the noise multiplier $\sigma$ in the added Gaussian noise $z\sim \mathcal{N}(0,C^2\sigma^2)$ and use the numerical accountant to obtain the final $\epsilon$'s.}
Experiments are repeated 5 times with different random seeds and we report the average.

\subsection{Private Training of Convolutional Neural Network from Scratch}
We conduct a standard image classification task using a wide Resnet \cite{zagoruyko2016wide}, WRN28-4 with 1.5 million parameters.
In contrast to the NLP fine-tuning tasks, we train the network from scratch.
The target dataset is CIFAR10 \cite{krizhevsky2009learning} containing 10 classes of real-world images.

\noindent\textbf{Baselines.}
We implement the same corresponding set of baselines described in Section \ref{subsec:nlp} for the CV task: non-private SGD (\textsf{N.P.}), vanilla DPSGD (\textsf{DPSGD}), RGP (\textsf{RGP}), and Sparse DPSGD (\textsf{Sparse DPSGD}).

\noindent\textbf{Experimental details.}
Our procedure of low-rank approximation and sparsification is applied to all convolutional layers.
The hyperparameters are set to the following, following \cite{rgp,zagoruyko2016wide}: Momentum SGD optimizer is used, clipping size 1 for \textsf{LSG}/\textsf{RGP} and 5 for \textsf{DPSGD}/\textsf{Sparse DPSGD}; batch size 1,000 for \textsf{LSG}/\textsf{RGP} and 350 for \textsf{DPSGD}/\textsf{Sparse DPSGD}\footnote{We are unable to run the experiment with a batch size of 1,000 due to memory constraints; it requires more than 32GB of GPU memory.}; epoch 200 for \textsf{LSG}/\textsf{RGP} and 100 for \textsf{DPSGD}/\textsf{Sparse DPSGD}; learning rate decayed from $0.5$ to $4 \cdot 10^{-3}$; the privacy parameter $\delta=10^{-5}$.
The three parameters we are interested in are $\epsilon$, $r$ and $p$.
We scan over these parameters ($\epsilon \in \{0.8, 1.7, 3.3, 6.8\}$, $r \in \{2, 4, 8, 16\}$, and $p \in \{0, 0.1, 0.3, 0.5\}$) to compare the performances.
Experiments are repeated 5 times with different random seeds and we report the average.

\subsection{Evaluation Results and Discussions}
\begin{figure}[ht]
     \centering
     \begin{tabular}{cc}
         \begin{subfigure}[b]{0.49\textwidth}
             \centering
             \includegraphics[width=\hsize]{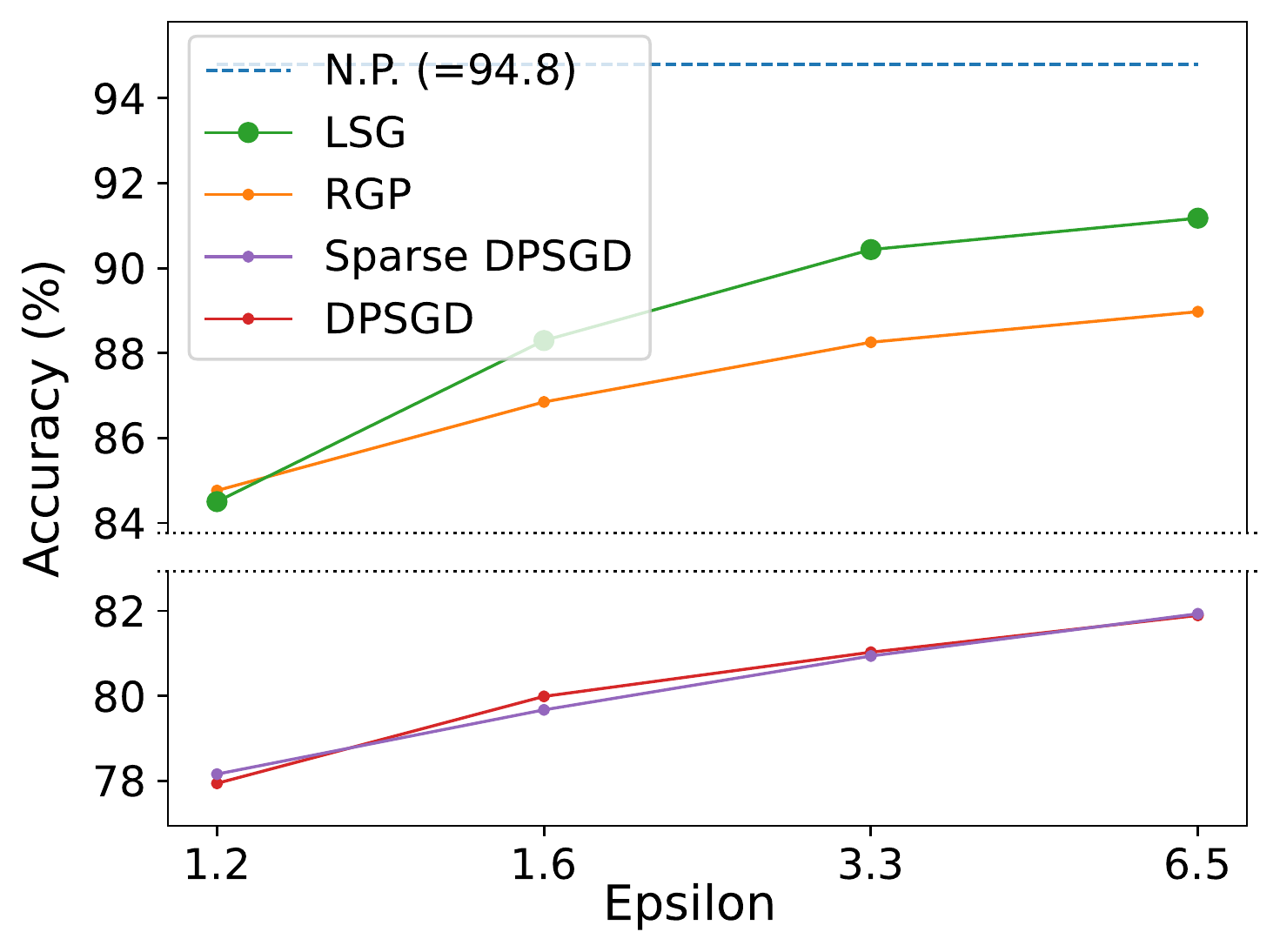}
             \caption{SST-2 \label{fig:exp_overall_sst2}}
         \end{subfigure}
         \begin{subfigure}[b]{0.49\textwidth}
             \centering
             \includegraphics[width=\hsize]{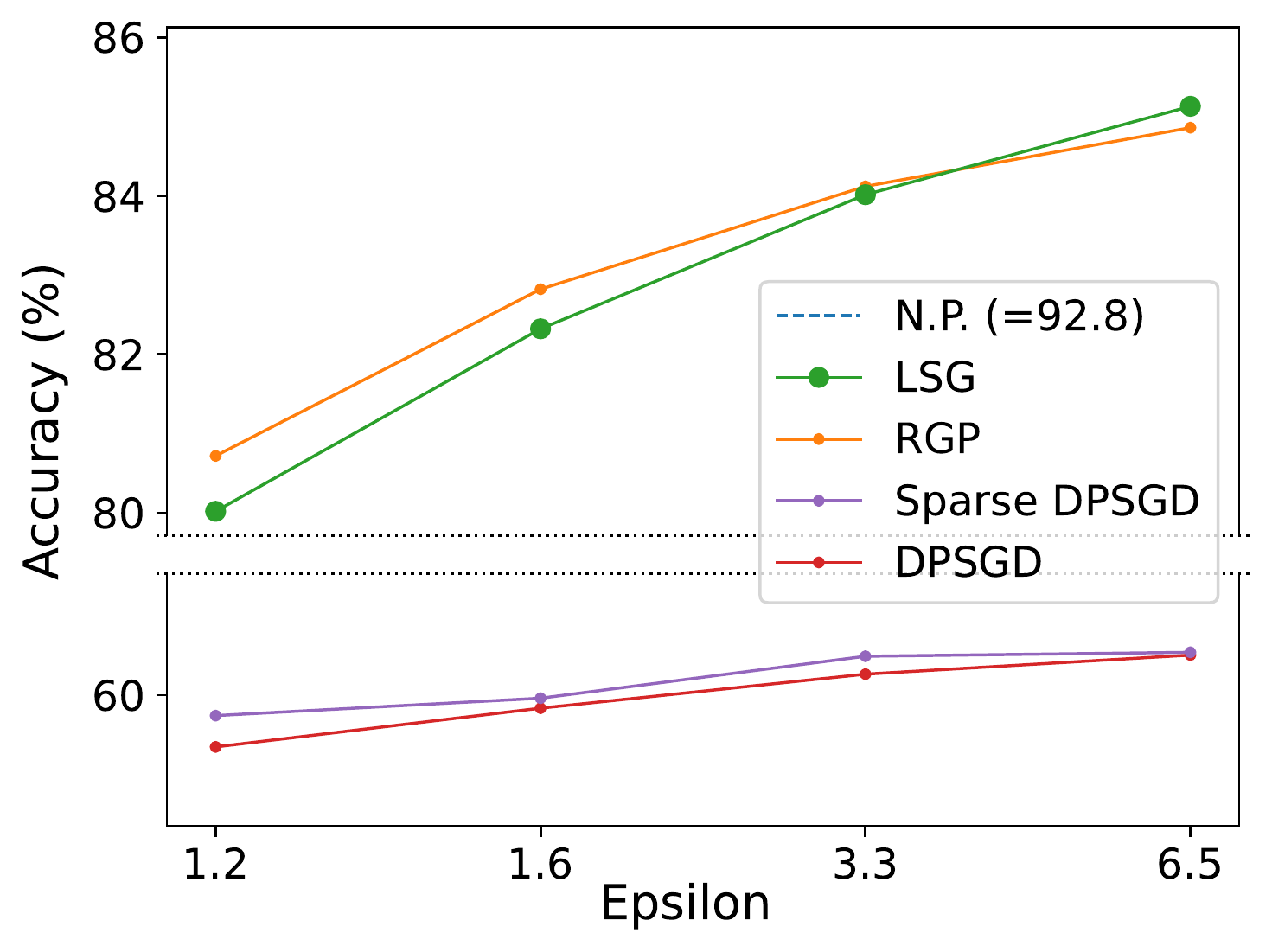}
             \caption{QNLI \label{fig:exp_overall_qnli}}
         \end{subfigure}
     \end{tabular}
     
     \begin{tabular}{cc}
         \begin{subfigure}[b]{0.49\textwidth}
             \centering
             \includegraphics[width=\hsize]{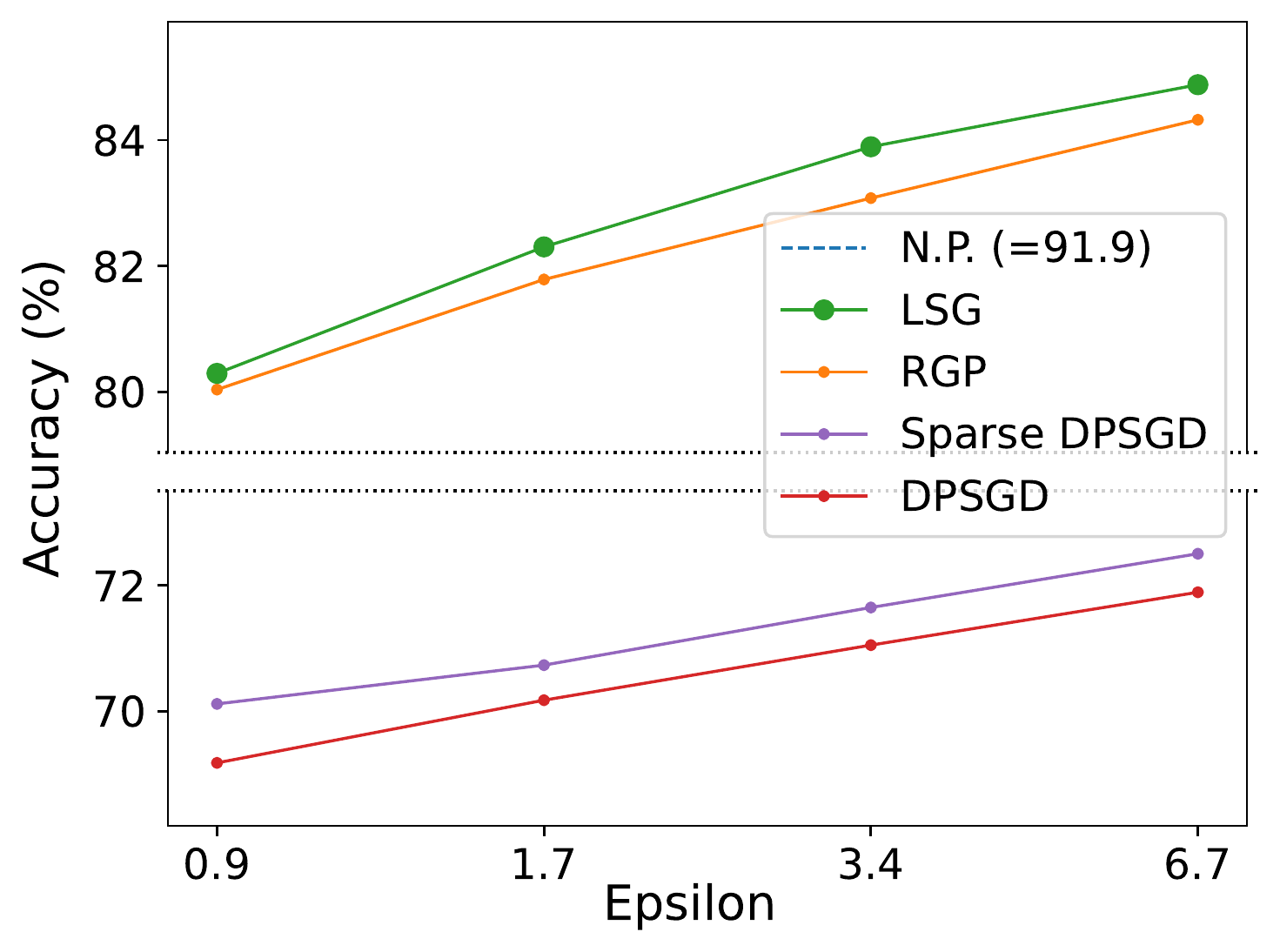}
             \caption{QQP \label{fig:exp_overall_qqp}}
         \end{subfigure}
         \begin{subfigure}[b]{0.49\textwidth}
             \centering
             \includegraphics[width=\hsize]{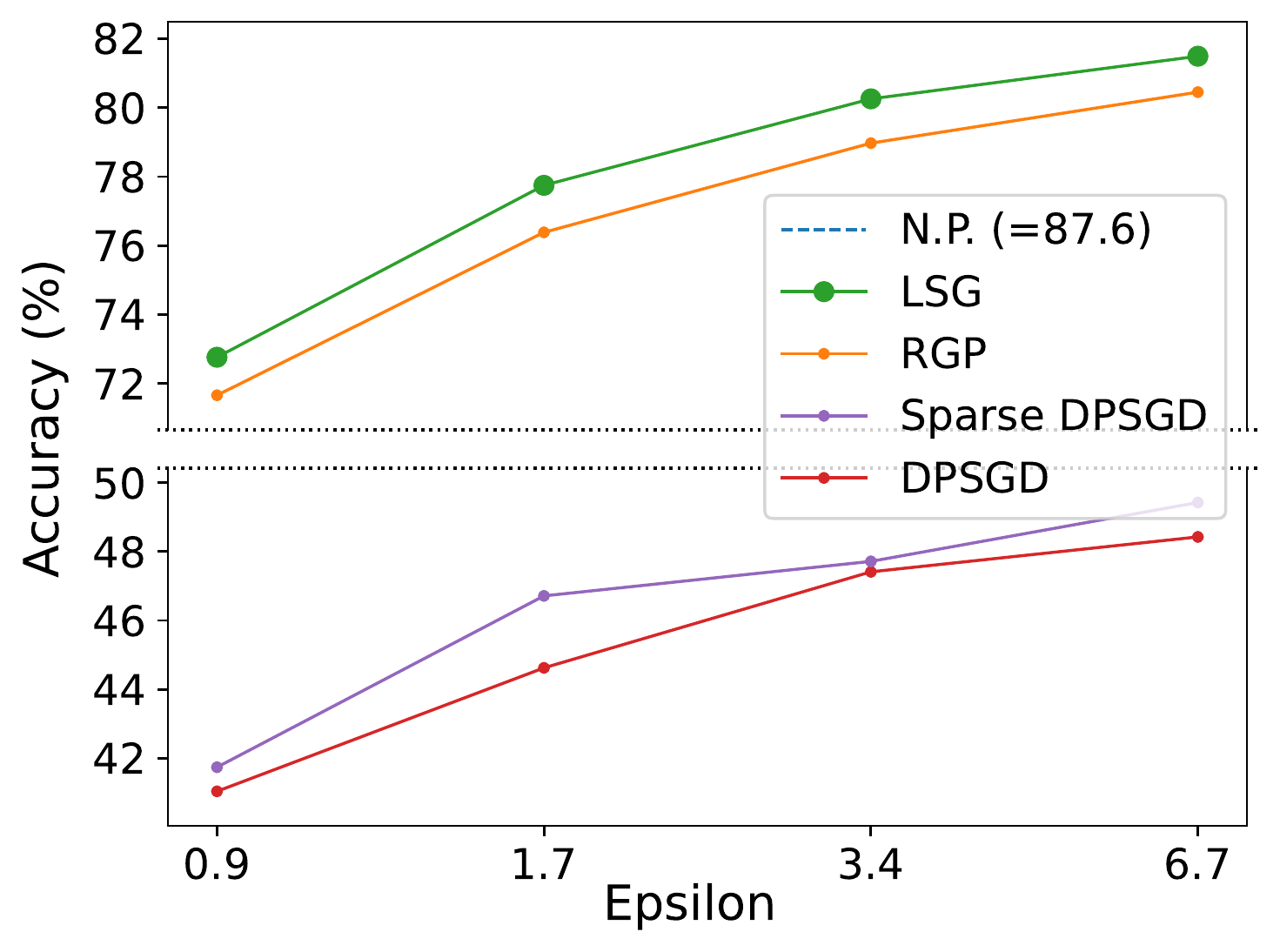}
             \caption{MNLI \label{fig:exp_overall_mnli}}
         \end{subfigure}
     \end{tabular}
     \caption{Accuracy (\%) of RoBRETa's downstream tasks with varying $\epsilon$. \textsf{N.P.} is quoted from the original work~\cite{roberta}. \label{fig:exp_overall_nlp}}
\end{figure}

\noindent\textbf{Results on NLP tasks.}
Figure \ref{fig:exp_overall_nlp} shows the overall results of DP fine-tuning, where plots of accuracy versus $\epsilon$ for each NLP task and fine-tuning method are shown.
Overall, \textsf{LSG} performs the best. 
As can be seen from the figure, there is almost always an improvement of up to 4\% compared to \textsf{RGP}, the best performing baseline.
While utilizing either of low-rank decomposition and sparsification for DP fine-tuning tends to improve the accuracy to a certain degree, combining both with \textsf{LSG} achieves the optimal performance. 
Our results also indicate that NN gradients inherently are low-rank and sparse; exploiting these properties helps mitigate the negative effects of clipping and noise.

Table \ref{tbl:exp_micro} shows the impact of rank $r$ and sparsity $p$ on DP fine-tuning performance for selected tasks and fixed $\epsilon$'s.
The smaller (higher) the value of $r$ ($p$) is, the lower-rank (sparser) the gradients are.
As before, the overall performance is improved by both low-rank decomposition and sparsification.
On the other hand, reducing the number of training parameters excessively (choosing the smallest $r$ and largest $p$ simultaneously) leads to performance deterioration, so there is a trade-off between utility and training parameter reduction.
It can also be observed that optimal performance is achieved only when both appropriate rank and sparsity are used.
Simply choosing the best $r$ and $p$ independently (e.g., choose the best $r$ with $p=0$, the best $p$ in the ``full" column of Table \ref{tbl:exp_micro}) and using \textsf{LSG} with these parameters does not lead to the best result.
This implies that there is certain interplay between low-rankness and sparsity, the two seemingly independent features of NN weights, that leads to a sweet spot of performance.
This warrants further (possibly theoretical) investigation and can be an interesting future direction.

\begin{table}[ht]
    \caption{Accuracy (\%) of RoBERTa's downstream tasks with varying rank $r$ and sparsity $p$. $p>0,r\leq8$ represents the proposal, \textsf{LSG}. $p=0,r\leq8$ represents the \textsf{RGP} baseline.
    The ``full" column represents \textsf{Sparse DPSGD}.
    Highest scores are in bold. \label{tbl:exp_micro}}
    \centering
    \begin{minipage}[t]{0.49\textwidth}
        \centering
        
        \subcaption{SST-2, $\epsilon=3.3$ \label{tbl:exp_micro_SST2}}
        \begin{tabular}{cr||rrrr}
            & & \multicolumn{4}{c}{Rank $r$} \\
            & & 2 & 4 & 8&full \\
            \hline \hline
        \multirow{4}{*}{\rotatebox[origin=c]{90}{Sparsity $p$}}
            & 0   & 85.4 & 87.9 & 88.3& 81.0 \\
            & 0.1 & 87.3 & 88.6 & 89.0&80.9 \\
            & 0.3 & 88.3 & 89.6 & 89.9&80.5 \\
            & 0.5 & 89.4 & 89.8 & \textbf{90.4}& 80.3\\
        \end{tabular}
    \end{minipage}
    \begin{minipage}[t]{0.49\textwidth}
        \centering
        \subcaption{QQP, $\epsilon=3.4$ \label{tbl:exp_micro_QQP}}
        \begin{tabular}{cr||rrrr}
            & & \multicolumn{4}{c}{Rank $r$} \\
            & & 2 & 4 & 8& full \\
            \hline \hline
        \multirow{4}{*}{\rotatebox[origin=c]{90}{Sparsity $p$}}
            & 0 & 82.8 & 82.8 & 83.1&71.1 \\
            & 0.1 & 83.3 & 83.3 & 83.3&71.2 \\
            & 0.3 & \textbf{83.9} & 83.7 & 83.8&71.7 \\
            & 0.5 & 83.8 & 83.6 & 83.6&71.5 \\
        \end{tabular}
    \end{minipage}
\end{table}

\noindent\textbf{Results on CV tasks.}
Table \ref{tbl:exp_cv} shows the results of DP training of models from scratch. For all $\epsilon$'s, our method outperforms not only \textsf{DPSGD} but also \textsf{Sparse DPSGD}, which uses only sparsity, and \textsf{RGP}, which uses only low-rankness of NNs. This again indicates that utilizing one of sparsity or low-rankness is not enough; both contributes to efficient DP training of NNs.

\begin{table}[ht]
    \caption{Accuracy (\%) of WRN28-4 on CIFAR10 with varying $\epsilon$. \textsf{N.P.} is quoted from the original work~\cite{rgp}. Highest scores for each $\epsilon$, except for \textsf{N.P.}, are in bold. \label{tbl:exp_cv}}
    \centering
    \begin{tabular}{l||r|r|r|r}
        Method & $\epsilon=0.8$ & $\epsilon=1.7$ & $\epsilon=3.3$ & $\epsilon=6.8$ \\
        \hline
        \textsf{LSG}         & \textbf{42.2} & \textbf{50.9} & \textbf{58.1} & \textbf{62.5} \\
        \textsf{RGP}          & 41.5 & 50.4 & 57.2 & 61.9 \\
        \textsf{Sparse DPSGD} & 23.4 & 35.2 & 48.4 & 60.2 \\
        \textsf{DPSGD}        & 23.2 & 35.1 & 48.4 & 59.9 \\
        \textsf{N.P.}         & \multicolumn{4}{c}{93.3}
    \end{tabular}
\end{table}

\begin{figure}[ht]
    \centering
    \includegraphics[width=0.9\hsize]{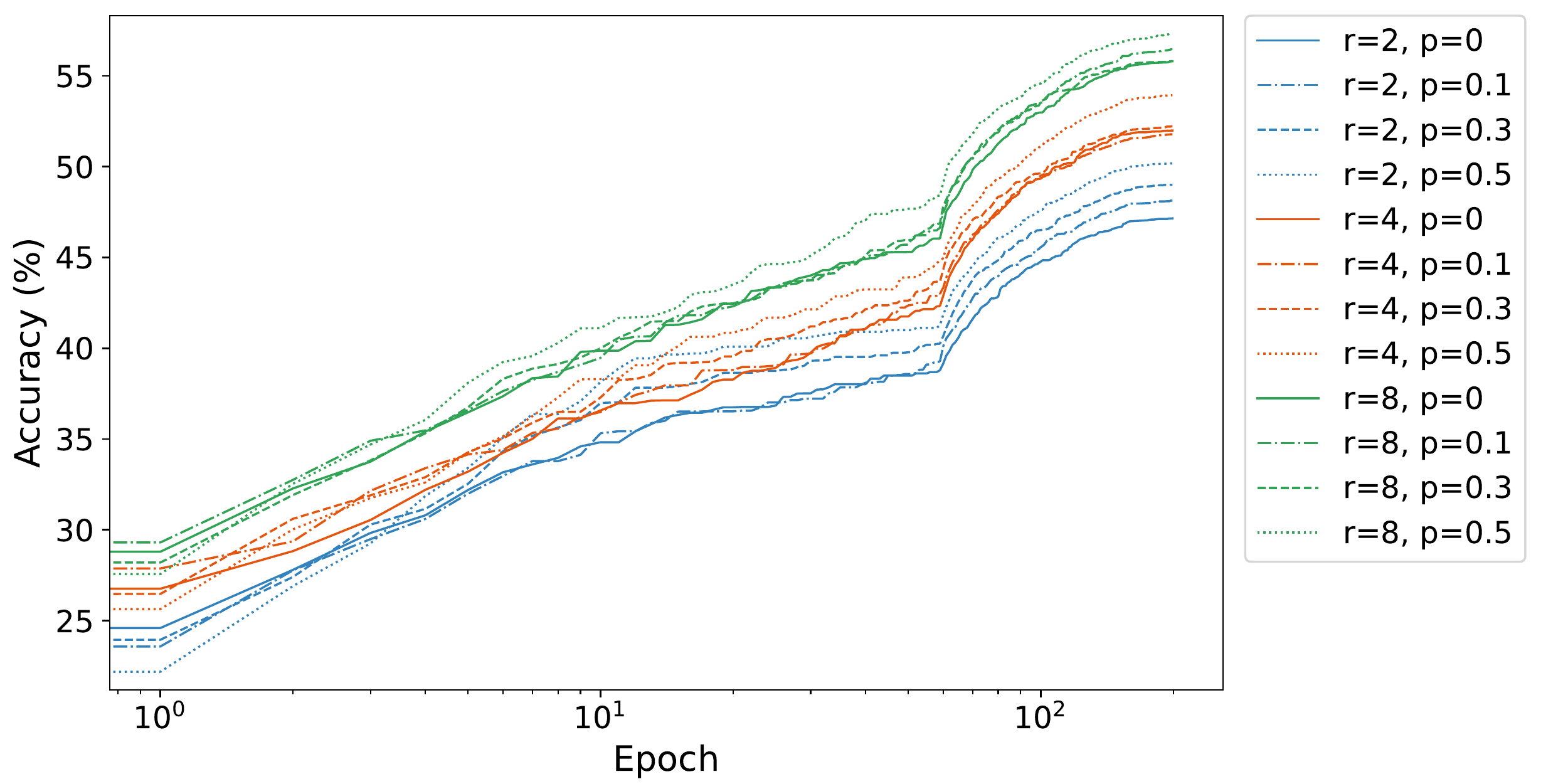}
    \caption{Accuracy (\%) of WRN28-4 on CIFAR10 ($\epsilon=3.3$) until 200 epoch with varying rank $r$ and sparsity $p$. $p>0$ represents \textsf{LSG} and $p=0$ represents \textsf{RGP}. The same color is a group of the same rank $r$, the same line style is a group of the same sparsity $p$. The finer dash, the greater sparsity. \label{fig:exp_convergence}}
\end{figure}

Figure \ref{fig:exp_convergence} shows the CIFAR10 learning curve of \textsf{LSG} over training epochs varying rank $r$ and sparsity $p$ (fixing $\epsilon = 3.3$). 
It is observed that in the early stage of training (up to about 5 epochs), training with sparsity ($p>0$) leads to sub-par performance. 
In later stages however, the accuracy improves and training with sparsity subsequently leads to better final performance. 
There are two reasons for this behavior.
First, as we perform sparsification based on the NN weights' importance, the initially randomized NN weights do not provide useful information for sparsification, and affect the training. 
Second, the learning rate is high in the early stage, further destabilizing the learning process.
It is interesting to consider ``switching on" sparsification only in the later stages of training to achieve better performance, which we leave for future work.
Note that the abrupt change of accuracy at around the 160-th epoch is due to the fact that the learning rate is changed dynamically and discontinuously at that epoch; it is not caused by our low-rank approximation or sparsification procedure.

\subsection{What did not Work}
\label{subsec:negative}
We have also explored approaches exploiting the NN redundancy alternative to the one proposed above.
Here, we report negative results that either have not fully utilized the redundant structure of NNs, or have not led to improvement in performance.

\noindent\textbf{Random sparsification.}
Instead of performing sparsification by NN weights' importance metrics, we have experimented with randomly sparsifying NN weights.
In particular, we have tested with 1) randomly sparsifying the low-rank gradient matrices, 2) randomly sparsifying the original gradient matrices (without low-rank approximation).
Both approaches did not improve the performance.
One explanation for this is that one cannot simply ``hit the lottery" \cite{lth} with random guessing (sparsification). 

\noindent \textbf{Sparsification based on low-rank matrices.}
Recall that our approach of performing sparsification on the low-rank matrices is based on the importance metrics of the original NN weights $\boldW$.
Another plausible way is performing sparsification on the low-rank matrices based on the importance of the low-rank matrices' weights themselves.
We did not find that this leads to performance improvement however.

\noindent\textbf{Direct sparsification.}
Yet another option might be to sparsify $\partial{\boldW}$ directly.
That is, one first uses the low-rank gradient matrices (sanitized with DP) to calculate $\partial{\boldW}$ with Equation \ref{eq:rgp} as performed in the first step of \textsf{LSG}.
Then, one directly sparsifies the resulting $\partial{\boldW}$.
However, this approach does not lead to an overall improvement because it cannot influence the DP-sanitized $\partial{\boldL}$ and $\partial{\boldR}$, and cannot mitigate the effects of clipping and noise imbued with them.

\section{Conclusion}
In this work, we have proposed a unified framework, named \textsf{LSG}, of training deep neural networks with differential privacy utilizing low-rankness and sparsity of neural networks.
\textsf{LSG} allows us to enjoy efficient DP deep learning with various network architectures and learning approaches, as confirmed by our thorough experimental evaluation.

\bibliographystyle{splncs04}
\bibliography{refs}

\end{document}